# Segmentation-free Vehicle License Plate Recognition using ConvNet-RNN

Teik Koon Cheang (*Author*)
Centre for Computing and Intelligent Systems
Universiti Tunku Abdul Rahman
Kajang, Malaysia
cheangtk@1utar.my

Yong Shean Chong
Centre for Computing and Intelligent Systems
Universiti Tunku Abdul Rahman
Kajang, Malaysia
yshean@1utar.my

Yong Haur Tay
Centre for Computing and Intelligent Systems
Universiti Tunku Abdul Rahman
Kajang, Malaysia
tayyh@utar.edu.my

*Abstract*— While vehicle license plate recognition (VLPR) is usually done with a sliding window approach, it can have limited performance on datasets with characters that are of variable width. This can be solved by hand-crafting algorithms to pre-scale the characters. While this approach can work fairly well, the recognizer is only aware of the pixels within each detector window, and fails to account for other contextual information that might be present in other parts of the image. A sliding window approach also requires training data in the form of pre-segmented characters, which can be more difficult to obtain. In this paper, we propose a unified ConvNet-RNN model to recognize real-world captured license plate photographs. By using a Convolutional Neural Network (ConvNet) to perform feature extraction and using a Recurrent Neural Network (RNN) for sequencing, we address the problem of sliding window approaches being unable to access the context of the entire image by feeding the entire image as input to the ConvNet. This has the added benefit of being able to perform end-to-end training of the entire model on labelled, full license plate images. Experimental results comparing the ConvNet-RNN architecture to a sliding window-based approach shows that the ConvNet-RNN architecture performs significantly better.

## I. INTRODUCTION

VLPR technology has seen increasing adoption over the past years [1]. This technology is useful to government agencies to track or detect stolen vehicles or for data collection purposes for traffic management improvement. Other applications for VLPR technology include parking lot management, automated ticket issuing and toll payment collection.

While VLPR is well studied, this task is still highly challenging and very often requires different modeling for each license plate arrangement, which varies across regions. This difficulty arises due to the wide variability in the visual appearance of text on account of different fonts, layouts and environmental factors such as lighting, shadows, and occlusions as well as image acquisition factors such as motion and focus blurs.

A common approach that extends CNNs to perform one-to-many problems is to transform it into a single-label classification problem by using allowing the CNN to only classify one character at a time. This gives rise to other problems, such as how to segmentate the characters that are passed to the CNN. [2] summarised two methods for character sequence recognition, the input segmentation (INSEG) method and the output segmentation (OUTSEG) method. More recent works by [3] combine OUTSEG with CNN-BRNN and CTC [4] to achieve state of the art results.

In this paper, we focus on recognizing license plate characters from real-world traffic cameras. The complexities faced by these cameras are mainly varying lighting conditions, motion blurs, variable scaling or skewing and translations due to imperfect bounding boxes from an imperfect localization algorithm. We propose a unified approach that integrates the segmentation and recognition steps via the use of a deep convolutional neural network and recurrent neural network (RNN) that operate directly on the image pixels. This model uses a combination of VGG-Net and a RNN, implemented using Torch, a well-developed deep learning framework.

We evaluated this approach on a vehicle license plate (VLP) dataset obtained from our industry partner and achieve over 76% accuracy in recognizing the license plate characters in our dataset, with a per-character accuracy of 95.1%.

The key contributions of this paper are: (a) a unified, end-to-end trainable model to localize, segment, and recognize multiple characters from real-world captured license plate photographs, and (b) results that show our method is comparable with hand-engineered methods despite its simplicity.

## II. RELATED WORK

### A. Optical Character Recognition (OCR)

Character sequence recognition is a widely researched topic and is intimately related to VLP recognition. Here we present a few techniques used by other researchers for word recognition and VLP recognition. [2] summarised two methods for character recognition, the input segmentation (INSEG) method and the output segmentation (OUTSEG) method.

In the INSEG approach, an algorithmic approach is used to determine the possible segmentation points of each character in a sequence. Only the image points containing a possible character are passed into the recognizer, and a sequence is generated by a concatenation of the most likely characters, selected through another algorithm such as Viterbi [2]. Some ways to determine the segmentation points for each character are connected component analysis (CCA) [5], Hough transform [6], and projection analysis [7].

For the OUTSEG approach, in contrast to INSEG, the candidate character segmentations are not selected beforehand. Instead, many tentative characters are generated by indiscriminately sweeping a small window over the entire image in small steps. At each step the contents of the window would be taken as a tentative character. Each tentative character is then passed into the recognizer, and the final sequence is decided after analysing the outputs of the recognizer as actual characters usually generate results that overlap significantly over many frames of the window [2]. In the work of [2], they decide the final output sequence by selecting characters with consecutive same characters and are segmented by a nil character.

INSEG and OUTSEG approaches both have the same problem where context of the full image is lost when only one frame of the entire image is passed into the recognizer. The recognizer is unable to obtain possibly useful information from previous or future frames.

*B. End-to-End Methods*

To counter this, instead of an INSEG or OUTSEG approach, there is a rise of using end-to-end methods in text recognition tasks [4, 8]. An end-to-end method describes a method that allows a sequence-to-sequence labelling where an input sequence is directly mapped onto an output sequence. Aside from avoiding the loss of context, this approach also helps significantly reduce the amount of heuristics in the system by allowing the Machine Learning models to directly model the entire process.

Another end-to-end method is connectionist temporal classification (CTC). CTC [4] primarily means that RNNs are trained to label unsegmented sequence data, and has been shown by them to perform better than framewise counterparts without CTC. The authors of [3] were able to use CTC in a LSTM-based recognition pipeline to perform the OUTSEG sequencing, achieving state of the art accuracy in VLP recognition.

Aside from CTC, there are other end-to-end approaches that use a CNN for feature extraction a RNN for sequencing or classification. In the works of [9], a Long-term Recurrent Convolutional Network (LRCN) is used to label or describe image sequences and describe static images. The LRCN model uses an LSTM as the sequencer to remember long range dependencies which are needed in video description tasks. [10] proposes a similar CNN-LSTM model to perform multiple label image classification. [11] uses a CNN-RNN model for RGB-D image classification task with a two CNNs for the RGB and Depth channels respectively, with outputs from both CNNs passed into multiple RNNs to perform the classification.

### III. METHOD

We propose a CNN-RNN model for VLP recognition problem. This framework is an end-to-end approach for VLP recognition with minimal modification required by allowing the model to do the recognition on entire VLP images without any prior segmentation or transformation.

*A. Problem Description*

In our experiments, we restrict our problem to only recognise typical Malaysian VLPs. A valid typical Malaysian VLP sequence has the following format:

$$S[C][C]N[C] \quad (1)$$

S = Character in the list {A,B,C,D,J,K,M,N,P,R,T,W,Z},
C = Character in the list

{A,B,C,D,E,F,G,H,J,K,L,M,N,P,Q,R,S,T,U,V,W,X,Y,Z},
N = Integer number from 1 to 9999.
Square brackets denote optional characters.

Because such end-to-end approaches require the output sequence length to be bounded, leading zeros were padded to the VLP labels in the dataset so that all labels are of the same length. For example, WLV3092 will be padded with three zeros to become 000WLV3092 so that it has a length of 10 characters. Very few license plate numbers contain more than eight characters, so we may assume the sequence length n is at most some constant N, with N = 10 sufficient for our experiments using Malaysian VLPs.

*B. Learning Features from VLP Images*

We attempt to train a variant of the 16-layer VGGNet [12] to output the full padded sequence of the VLP, given input images without any preprocessing, except for resizing to a standard size of $240 \times 120$ pixels. A batch normalization process is added after each convolutional layer, as it is shown to be able to speed up training as well as boosting performance. This can be visualised as a separate batch normalization layer after every convolution layer.

For a usual multi-class classification task, the number of the final layer would be trained to output a probability distribution of length C, where each probability pc corresponds to the probability of the class label. However, we modify this to create a sequence labelling approach by allowing the network to directly output the entire sequence in one forward pass of the input image. An architecture similar to [13] is used, where each character in the sequence is represented by C neurons in the output layer, resulting in a total of $K \times C$ number of neurons in the output layer. We assign the first C number of output neurons to the first character of the sequence, the second C number of neurons to the second character and so on.

Our approach allows the hidden units of CNN to have access to the features in the entire image, instead of only seeing one small window image when using a sliding window approach.

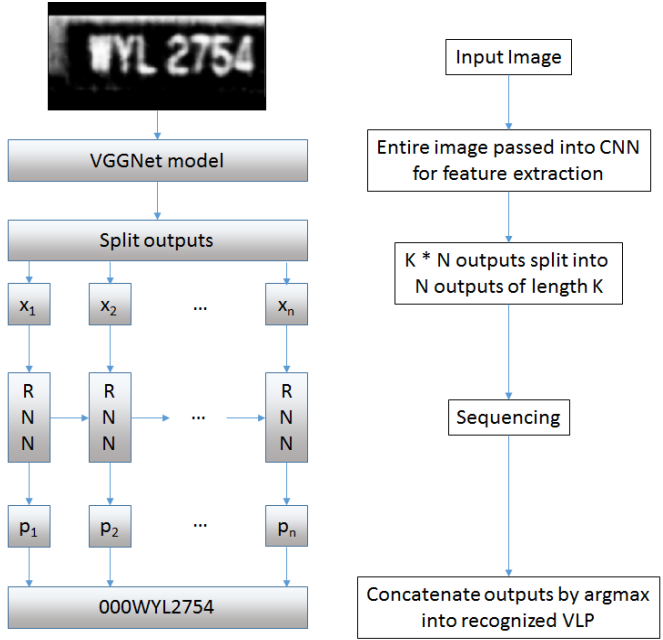

Fig. 1. CNN-RNN Architecture. The output of CNN becomes the input of RNN, which enables RNN to learn the sequential order of character features.

*C. Learning Sequence of Features*

In the CNN-RNN architecture (as shown in Fig. 1), the outputs from the CNN are not directly used as results. Instead, they are passed into a RNN with 36 hidden units for further sequencing. We then sample K outputs from the RNN orderly, with the argmax of the k-th output being the k-th character of the VLP. The recurrent layer takes the label embedding of the previously predicted label, and models the co-occurrence dependencies in its hidden recurrent states by learning nonlinear functions:

$$r(t) = h[r(t-1), w_k(t)] \quad (2)$$

$$o(t) = h[r(t-1), w_k(t)] \quad (3)$$

where $h(\cdot)$ is a transformation function (i.e., sigmoid), $r(t)$ and $o(t)$ are the hidden states and outputs of the recurrent layer at the time step $t$, respectively, $w_k(t)$ is the label embedding of the $t$-th label in the prediction path.

*D. Training*

Training of the CNN and CNN-RNN model is done by using the cross-entropy loss on the softmax normalization of score and back-propagation through time (for the CNN-RNN model) with stochastic gradient descent algorithm. Although it is possible to fine-tune the VGGNet in our architecture, we keep the network unchanged in our implementation for simplicity.

*E. Inference*

A prediction path is a sequence of labels $L = (l_1, l_2, ..., l_k, ..., l_K)$ with each $l_k = (p_1, p_2, ..., p_n, ..., p_C)$, where the probability of each label $l_t$ can be computed with the information of image I and the previously predicted labels $l_1, ..., l_{t-1}$. The RNN model predicts a sequence of labels by finding the prediction path that maximizes the a priori probability.

IV. EXPERIMENT

In our experiments, we resize all input VLP images to 240×120 pixels. The CNN module is the 16-layer VGG network without any pre-training. The dimension of the label embedding of the CNN is 360. We employ a decreasing learning rate of $0.1/(10 \times epoch)$ for all layers. For RNN module, the input dimension is the same as the output dimension of CNN. Our RNN module has only one hidden layer with 36 hidden units. We evaluate the proposed method on our private VLP dataset. The evaluation demonstrates that the proposed method achieves a comparable performance.

*A. Evaluation Criteria*

One way to evaluate the performance of a sequence labelling model is to measure the percentage of labels that are perfectly predicted by the model. However, only considering perfect predictions would not take into account the performance of the model in predicting individual characters. Therefore we also include other measures, namely, average edit distance and average ratio between two sequences. Edit distance is the number of changes required to change one sequence of characters into another. The allowed edits are insertion, deletion, and substitution. The ratio of two sequences are obtained by the formula:

$$((M + N) - distance)/(M + N) \quad (4)$$

where M and N is the length of each sequence to be compared, and distance refers to the edit distance. This is to take into account the difference between sequence lengths of both predicted and target sequences.

*B. Dataset*

VLP datasets are difficult to obtain due to the identifiable nature of VLP and privacy concerns. Therefore, there is no publicly available benchmarks and we will not publicize this dataset but include only sample images for visualization purpose. This dataset contains 2,713 labelled license plate images captured by a VLPR camera deployed in a real-world setting. 409 images are extracted from the dataset to be used for validation. This dataset also includes some noisy and incomplete license plate images, some are shown in the sample images in Fig. 2.

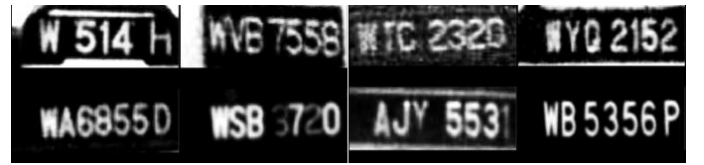

Fig. 2. Example of images in the dataset

## C. Results

TABLE I. THE EVALUATION RESULTS ON OUR VLP DATASET

| Setup | Percentage Perfect [a] | Average Edit Distance [a] | Average Ratio [a] |
|---|---|---|---|
| CNN only (no RNN) | 23.34% | 2.86 | 0.79 |
| CNN only, with data augmentation | 26.19% | 2.59 | 0.81 |
| CNN-RNN | 69.68% | 0.89 | 0.93 |
| CNN-RNN, with data augmentation | 76.53% | 0.74 | 0.94 |

[a.] A good method will have higher percentage perfect and ratio with lower edit distance

We show the results of different architecture setup in Table I. Augmentation used include random scaling, translation, blurring, sharpening and rotation.

The results show an incredible improvement of performance with the addition of RNN. This is because for non-segmented VLP images, the sequence of the characters becomes significant.

The VGGNet model uses multiple small convolutions with a kernel size of $3 \times 3$, hence it is difficult to discern the features learned by the CNN. However, it can be seen in Figure X that some of the filters vaguely resemble edge and point detectors.

From the confusion matrix in Fig. 3, we can see that most of the misclassifcations come from the character portion of the sequence, while the digits portion of the sequence has a much higher accuracy. This is likely due to larger number of digits in the VLP sequence and less number of classes for digits compared to the characters, which leads to the model having more exposure to varying digits compared to characters.

## V. CONCLUSION

The proposed CNN-RNN framework for vehicle license plate recognition combines the advantages of feature learning and joint image/label embedding by implementing CNN and RNN to model the feature and label sequence. Experimental results on Malaysian VLP dataset demonstrate that the proposed approach achieves comparable performance to the manually engineered methods such as sliding window method. Sliding window solutions require data that is labelled character by character, which is much harder to acquire, whereas our method only require one label for each VLP image, which is easier to obtain. We also show that the architecture can produce reliable results despite varying lighting conditions, blurring and noise.

However, recognizing VLPs with large translations is still challenging due to the majority of our dataset being reasonably centered. Also, as seen in the confusion matrix in Fig. 3, similar characters pairs such as M and N, D and Q, T and Y as well as C and G are frequently the cause of mis-recognitions.

Theoretically, the RNN should be able to model the valid sequences of the possible Malaysian VLP formats, as evidenced from the error rate on character pairs D and 0 from Fig 3 whereby there is very little confusion on this extremely similar pair of characters. The RNN, with the context of previous outputs, is able to recognize that it is invalid to have the other character at that position. However in practice it might not be possible to train the model with an exhaustive list of all possible VLP sequences.

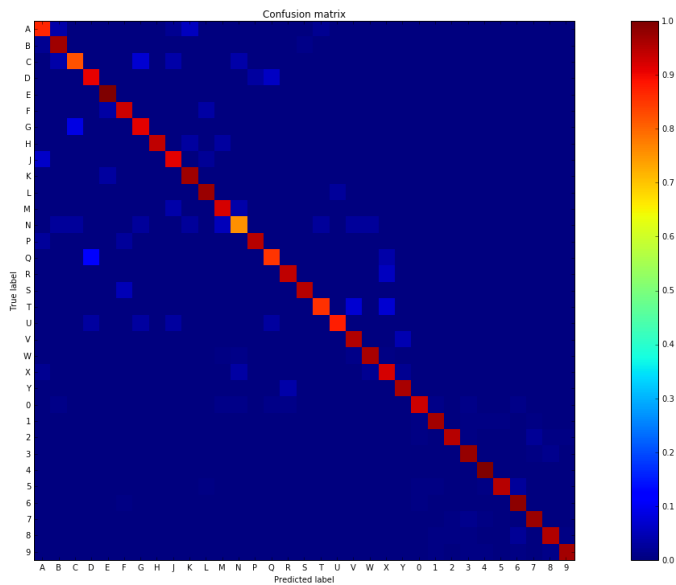

Fig. 3. Character-level confusion matrix, normalized to percentages. Note that results have been excluded in cases where predicted sequence length is different from actual sequence length. The char-level accuracy from this confusion matrix is 95.1%.

Hence for real applications, adding a lexicon to validate the VLP sequence would help obtaining results that conform to the Malaysian VLP format.

Replacing the RNN module with a long short term memory (LSTM) module could possibly improve the performance by being able to remember long term dependencies [3, 14, 15]. Also, more types of augmentation techniques could be used to improve the generalizability of the model such as Gaussian noise as well as extreme translations and skewing which may simulate the noise and translations present in the real-world data. We will investigate that in our future work.